\documentclass[conference]{IEEEtran}
\IEEEoverridecommandlockouts
\usepackage{cite}
\usepackage{balance} 
\usepackage{amsmath,amssymb,amsfonts}
\usepackage{algorithmic}
\usepackage{graphicx}

\usepackage{textcomp}
\usepackage{xcolor}
\usepackage{amsmath,amssymb,amsfonts}
\usepackage[ruled,linesnumbered,commentsnumbered]{algorithm2e}
\usepackage{multicol}
\usepackage{algorithmic}
\usepackage{graphicx}
\usepackage{tabularx}
\usepackage{textcomp}
\usepackage{xcolor}
\usepackage{float}
\usepackage[super]{nth}
\usepackage{subcaption}
\usepackage{booktabs}
\usepackage{multirow}
\usepackage{siunitx}
\usepackage{ragged2e}
\usepackage{threeparttable}
\usepackage{array}
\usepackage[super]{nth}
\usepackage{arydshln} 
\usepackage{amsmath,amssymb,amsfonts}
\usepackage{algorithmic}
\usepackage{graphicx}
\usepackage{textcomp}
\usepackage{xcolor}
\def\BibTeX{{\rm B\kern-.05em{\sc i\kern-.025em b}\kern-.08em
    T\kern-.1667em\lower.7ex\hbox{E}\kern-.125emX}}

\renewcommand{\baselinestretch}{1}

\begin{document}

\title{ Pelican: A Deep Residual Network for Network Intrusion Detection\\
}

\author{\IEEEauthorblockN{Peilun Wu\IEEEauthorrefmark{1}, Hui Guo\IEEEauthorrefmark{2} and Nour Moustafa\IEEEauthorrefmark{3}}
\IEEEauthorblockA{School of Computer Science and Engineering, University of New South Wales, Sydney \IEEEauthorrefmark{1}\IEEEauthorrefmark{2}\\
Australian Center for Cyber, University of New South Wales, Canberra\IEEEauthorrefmark{3}\\
Innovation Institute, Sangfor Technologies Inc.\IEEEauthorrefmark{1}\\
Email: \IEEEauthorrefmark{1}z5100023@zmail.unsw.edu.au,
\IEEEauthorrefmark{2}h.guo@unsw.edu.au, \IEEEauthorrefmark{3}nour.moustafa@unsw.edu.au}}

\maketitle

\begin{abstract}
One challenge for building a secure network communication environment is how to effectively discover and prevent malicious network behaviours.
The abnormal network activities threaten users' privacy and potentially damage the function and infrastructure of the whole network.
To address this problem, the network intrusion detection system (NIDS) has been used. By continuously monitoring network activities, the system can timely identify attacks and prompt counter-attack actions. 
NIDS has been evolving over years. 
The current-generation NIDS incorporates machine learning (ML) as the core technology in order to improve the detection performance on novel attacks. 
However, the high detection rate achieved by a traditional ML-based detection method is often accompanied by large false-alarms, which greatly affects its overall performance.
In this paper, we propose a deep neural network, Pelican, that is built upon specially-designed residual blocks.
We evaluated Pelican on two network traffic datasets, NSL-KDD and UNSW-NB15.
Our experiments show that Pelican can achieve a high attack detection performance while keeping a much low false alarm rate when compared with a set of up-to-date machine learning based designs.

\end{abstract}


\section{Introduction}
\begin{figure}[t]
    \centering
    \includegraphics[width=.75\linewidth]{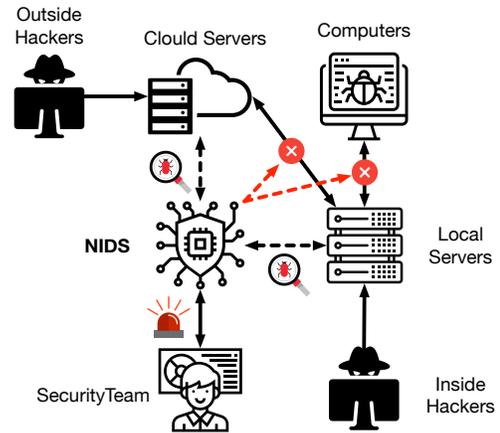}
    \caption{Network Intrusion Detection with NIDS}
    \label{fig:NIDS}
\end{figure}

With the continuously expanding network scale, network attacks are becoming more and more frequent, volatile and advanced.
From mobile phones, personal computers to industrial servers, all networked devices are potentially under the threats of malicious intrusion activities.
How to effectively discover and prevent the intrusions is very challenging.
Many cybersecurity enterprises and research institutes have been developing the network intrusion detection system (NIDS) to safeguard their networked computing environments. 
Fig.~\ref{fig:NIDS} shows a general setting of using such a detection system, where NIDS sits within the network, continuously monitors in-out network traffic, and reports any suspicious behaviours to the security  team for further attack identification and containment. 

The NIDS designs were initially signature-based and were only effective for the detection of known attacks. With the increasingly large traffic data generated each day and the need to timely detect volatile and advanced attacks (due to growing network users), the artificial intelligence (AI) or machine learning (ML) technology has come into play in the NIDS design. 	Most of the existing ML-based NIDS designs, however, achieve high detection rate at the cost of large false alarms, which is inevitably adding unnecessary workload to the security team and may delay the counter-attack responses, and hence adversely affecting the overall network security. 

\begin{figure*}[t]
    \centering
    
    \begin{subfigure}[b]{.45\linewidth}
        \centering
        \includegraphics[width=\linewidth]{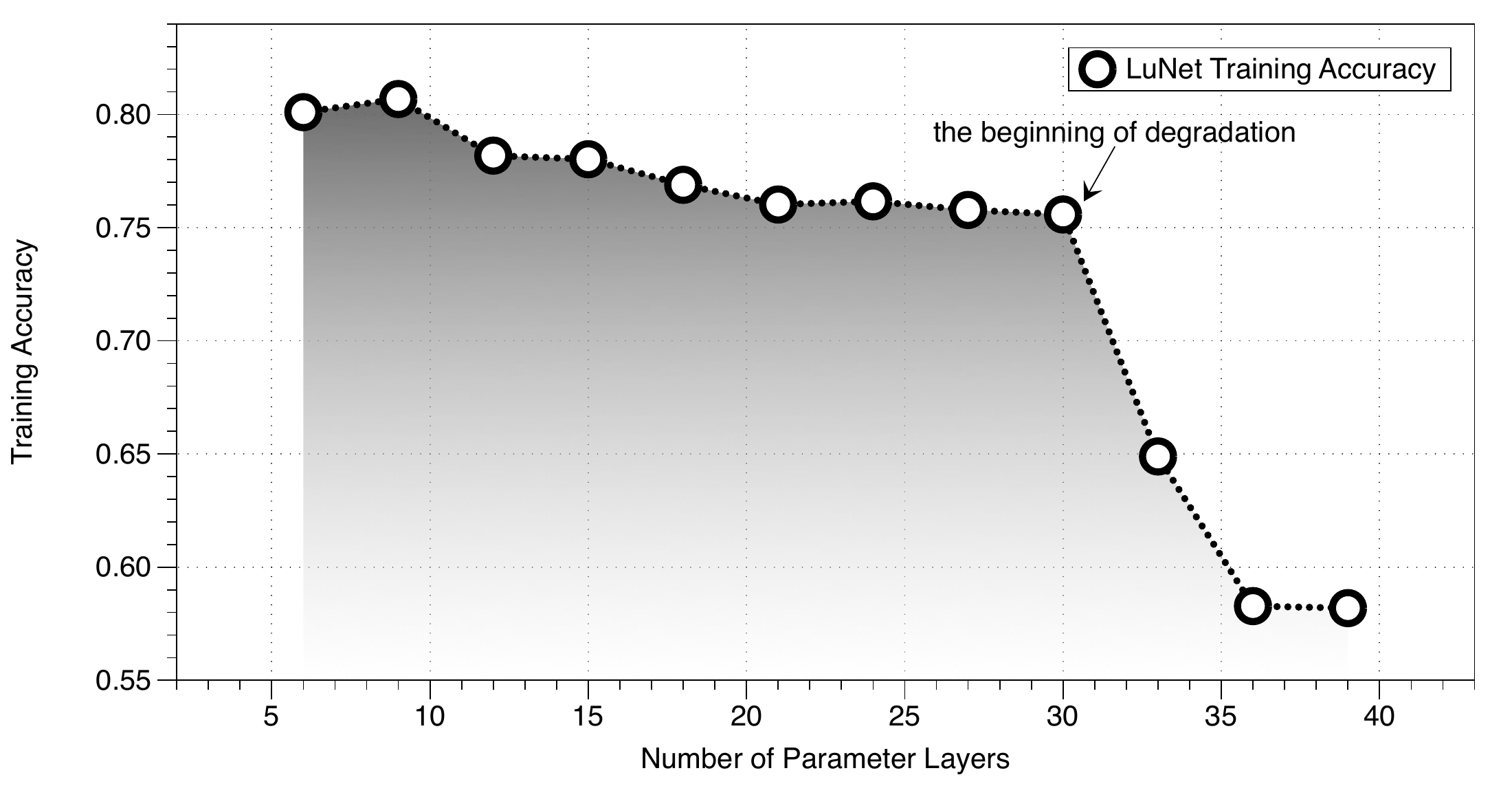}
        \caption{Training Accuracy of LuNet on UNSW-NB15}
    \end{subfigure}%
~
    \begin{subfigure}[b]{.45\linewidth}
        \centering
        \includegraphics[width=\linewidth]{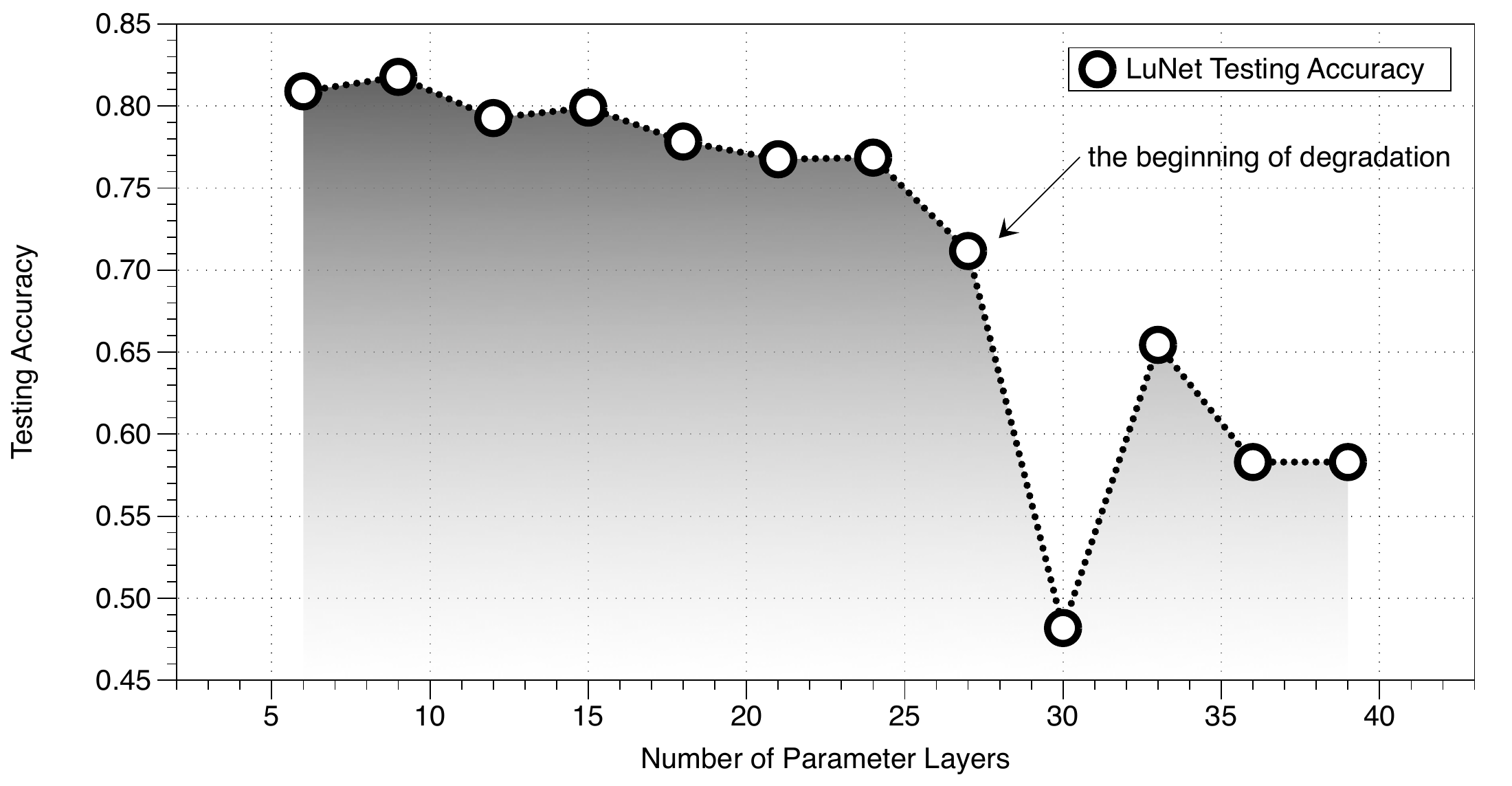}
        \caption{Testing Accuracy of LuNet on UNSW-NB15}
    \end{subfigure}
        \caption{Motivational Example: Performance Degradation in Training DNN for Network Intrusion Detection}
\label{fig.motivation}
\end{figure*}

In this paper, we aim to develop a deep neural network (DNN) design to achieve high intrusion detection accuracy. Our main contributions are as follows:
\begin{itemize}
    \item We investigate the performance degradation -- the hurdle existing in training DNN, and introduce residual learning in our design. 
    \item We propose a building-block based DNN structure, Pelican, 
    that is built on sub residual networks. 
    \item To improve learning efficiency, we incorporate CNN and RNN in the sub residual network so that both spatial and temporal features in the input data can be effectively captured.
    \item We test Pelican on two different network traffic datasets and compare Pelican with a set of ML based designs for intrusion detection. Our experiment results show that Pelican outperforms those existing designs. 
    \item Importantly, our work
    demonstrates that the residual learning is effective in building DNN for network intrusion detection. 
\end{itemize}

The motivation of using residual learning and the brief discussion of residual learning are given in the next two sections (Section~\ref{why} and Section~\ref{how}). Our design for Pelican with residual networks is presented in Section~\ref{pelican}.

\section{Motivation}
\label{why}

It is a general perspective that a deeper neural network should have a better potential on learning and generalization data than a shallow one.
We tested this idea by running an experiment based on an existing neural network design, LuNet, which was proposed in \cite{peilun} for network intrusion detection.
Fig.~\ref{fig.motivation} shows the plots of training accuracy and testing accuracy (on a dataset UNSW-NB15) of the network with respect to different number of parameter (or learning) layers. From these plots, we can see that as the network depth increases, the learning accuracy does not increase as expected; Instead the performance is even degraded. 

In fact, this {\bf performance degradation} in training deep neural network is not just for network intrusion detection. The problem was also revealed early in \cite{he2016deep,he2015convolutional,srivastava2015highway} when training the convolution neural networks (CNN) for image classification.

The performance degradation issue imposes a great hurdle in unleashing the potential of deep neural network. There have been a few solutions proposed. Among them, residual learning has been demonstrated an effective technique in the area of image classification. Here we want to testify that it is also effective for network intrusion detection. 

The general design idea of residual learning is briefly presented below.
(The detailed discussions can be found in an abundant literature, \cite{he2016deep} etc.) 

\section{Residual Learning}\label{how}

\begin{figure}[t]
\centering
\centerline{\includegraphics [width=.9\linewidth]{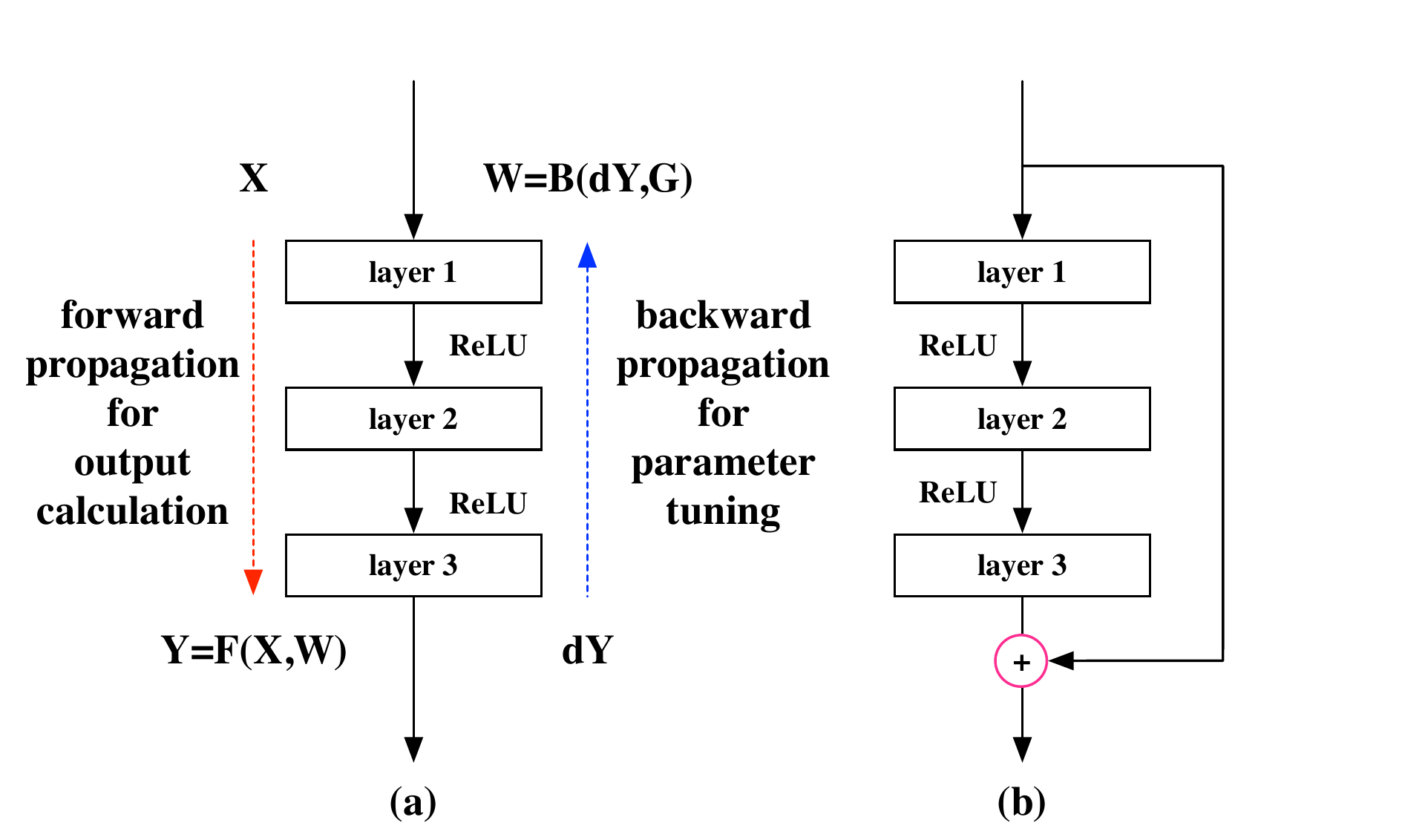}}
\caption{(a) Plain Network (b) Residual Network}
\label{fig.residual_learning_block}
\end{figure}

Residual learning is related to deep neural network. A deep neural network can be abstracted with multiple learning layers, as demonstrated in 
Fig. ~\ref{fig.residual_learning_block} (a), where three learning layers are used. Each layer in the network contains a  set of connections between its input and output, and has corresponding adaptive parameters (typically connection weights). 

Network training consists of rounds of rounds of learning that basically includes forward mapping and backward tuning. 

\textbf{Forward Mapping:} 
For each round of learning, the network processes the input data layer by layer in a forward-propagation fashion to generate an output mapping \\$F(X, W): X \rightarrow Y$, where $X$ and $Y$ are, respectively, the input and output vectors, and $W$, the total set of adaptive parameters from all learning layers. 
The learning error $dY$ generated at the output is then used to train the network. 

\textbf{Backward Tuning:} 
Basically, training a network is tuning the adaptive parameters in each learning round to minimize the learning error. There are several classes of training algorithms. 
For DNNs with a large number of adaptive parameters, a gradient descent optimization algorithm such as SGD, RMSprop, ADAELTA, is often used.

With such an algorithm, the error $dY$ is backward propagated through each layer where each of the related parameters, $w_i$, $w_i\in W$, is adjusted based on its gradient $g_i$, as shown in the below formula:
\begin{equation}
    w_i = w_i -r_i*g_i,
\end{equation}
where $r_i$ is the step size of the tuning and can be determined by the learning rate. 
We use $W = B(dY, G)$ to denote this tuning process in Fig. ~\ref{fig.residual_learning_block} (a), where $G$ is the total set of gradients.

The gradient, $g_i$, of $w_i$ is the product of a chain of partial derivatives of local outputs along the backward chain from the output layer to the current parameter layer related to $w_i$, and can be approximately denoted as  
\begin{equation}
    g_i  \approx \prod w_{j}. 
\end{equation}

If those parameters are smaller than 1 (as often is the case), the gradient tends to be vanished towards the initial learning layers; Therefore, the parameters of the initial layers will not get properly tuned in each training round. But the initial layers directly learn the input data and their learning is often critical in capturing the fundamental features of the input data and plays an import role for the final learning outcome.
On the other hand, if those parameter are larger than 1, the gradient will become exponentially large (namely, exploded); In this case, the parameter tuning is too rough and may fail to make any improvement in the training, and therefore, the learning cannot converge. 

Consequently, with more and more learning layers added, this problem become more and more eminent, leading to network learning performance decreased, as observed in our experiment shown in Section~\ref{why}. 
Use of residual learning can mitigate such a problem. 
Residual learning was first introduced in ResNet \cite{he2016deep}, a very deep neural network with hundreds of learning layers, for computer vision tasks. 
ResNet has demonstrated amazing results that are even better than human performance for image recognition.
Residual learning applies a short cut between the output layer to an input layer, as demonstrated in Fig.~\ref{fig.residual_learning_block}(b), so that the output error can be propagated to the input layer through a shorten route, hence avoiding the gradient vanishing/exploding problem caused by the existing long propagation path.

\section{Pelican}\label{Pelican}
\label{pelican}



\begin{figure}[t]
\centering
\centerline{\includegraphics [width=\linewidth]{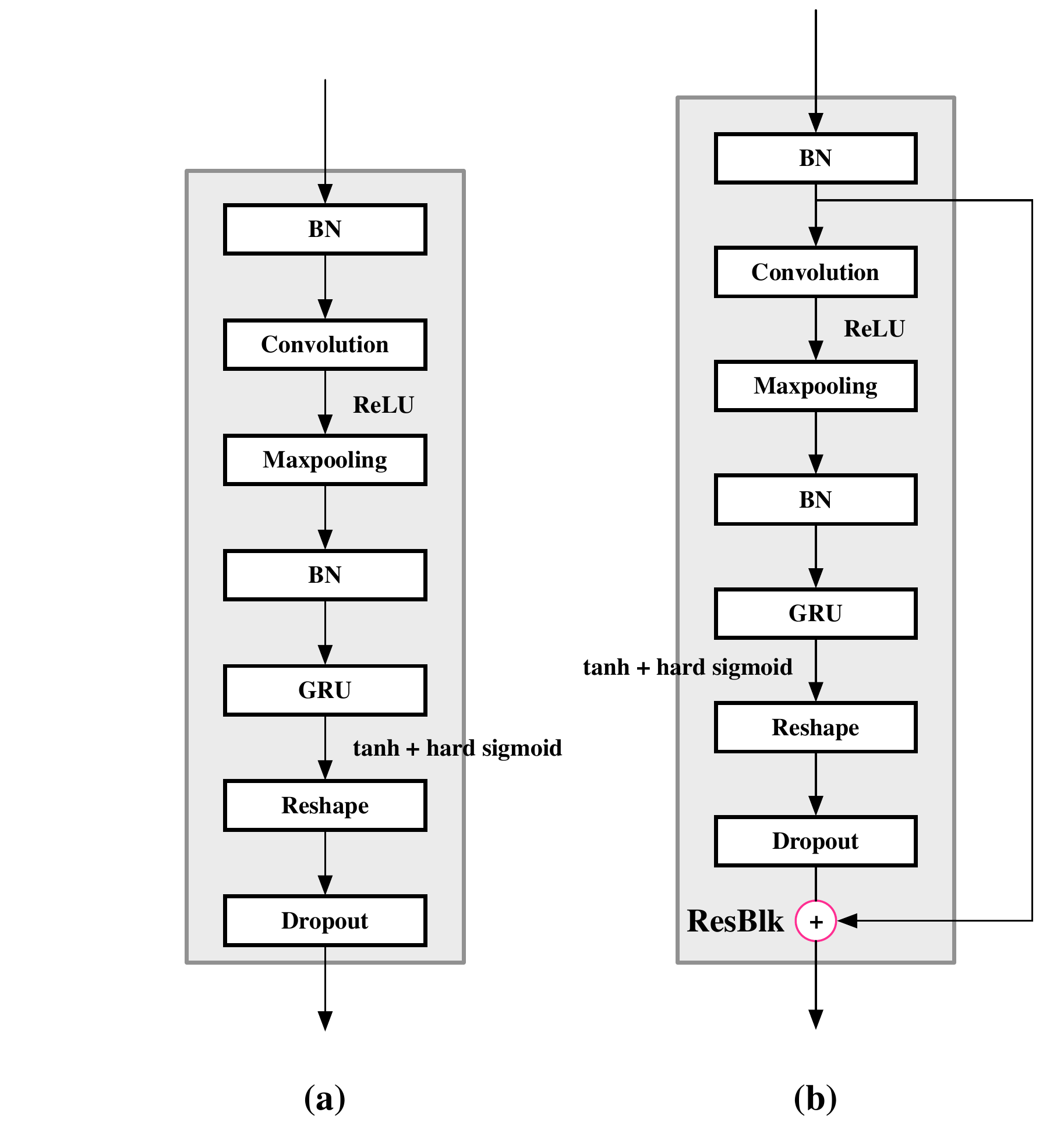}}
\caption{(a) Plain Block (b) Residual Block}
\label{fig.Peilican plain_residual_pelican}
\end{figure}

Pelican is basically constructed with a set of residual blocks, where each residual block, $ResBlk$, is a sub residual network that is formed based on a plain network block proposed in \cite{peilun}, as is shown in Fig.~\ref{fig.Peilican plain_residual_pelican}(a). The plain network block uses CNN and RNN and is able to learn both spatial and temporal features from the input. The main functions of each type of layer in the basic block are briefly summarized below.

\begin{itemize}
\item \textbf{Batch-Normalization (BN)} \cite{szegedy2015going}: BN reduces the internal covariate shift during training by scaling weights to unit norms. It is applied before Convolution and Recurrent network (GRU) layers.
BN helps fine-tune the learning rate to accelerate network training. 
\item \textbf{Convolution} \cite{krizhevsky2012imagenet}: The convolution operation in this layer extracts the spatial features from the input data 
and produces a feature map at the output. The convolution operation is often followed by an activation function to amplify distinctions between features generated by the convolution. Here the rectified linear unit (ReLU)  is used as the activation function.
\item \textbf{Maxpooling} \cite{zhou1988computation}: This layer selects most active neurons based on the maximum probabilities in nearby features to facilitate the next stage learning.
\item \textbf{Gated Recurrent Unit (GRU)} \cite{chung2015gated,chung2014empirical}: 
GRU is a recurrent network that can extract the temporal features of the input data through a recurrent process. 
Similar to the convolution layer, an activation function and a recurrent activation function  are needed for GRU, for which
tanh and hard sigmoid are, respectively, used here. 
\item \textbf{Reshape}: Since the dimension of data changes during learning, the reshape layer is used to keep the accordance of data dimension.
\item \textbf{Dropout} \cite{srivastava2014dropout}: This layer randomly drops out some connections from the network to prevent the network from overfitting. It must be noted that dropout is not a sole solution to overfitting, which will be further discussed in Section~\ref{limitation}.
\end{itemize}

With the functional layers presented above for the basic plain block, we construct 
the residual network block (\textit{ResBlk}), as shown in Fig.~\ref{fig.Peilican plain_residual_pelican}(b), where the short cut is connected from the BN output to facilitate the initialization of overall deep network.  
To investigate the effectiveness of residual network, we constructed two plain networks and two residual networks with different depths.
The brief description of the four networks is given below:

\begin{itemize}
    \item  \textbf{21 parameter-layer plain network (Plain-21 )}: It was built with five plain blocks + one global average pooling layer + one dense layer.
    \item  \textbf{21 parameter-layer residual network (Residual-21)}: It was built with five residual blocks + one global average pooling layer + one dense layer.
    \item  \textbf{41 parameter-layer plain network (Plain-41)}: It was built with ten plain blocks + one global average pooling layer + one dense layer.
    \item  \textbf{41 parameter-layer residual network (Residual-41 (Pelican))}: It was built with ten residual blocks + one global average pooling layer + one dense layer.
\end{itemize}

\begin{figure}[t]
\begin{subfigure}{\linewidth}
\centerline{\includegraphics[width=.95\linewidth]{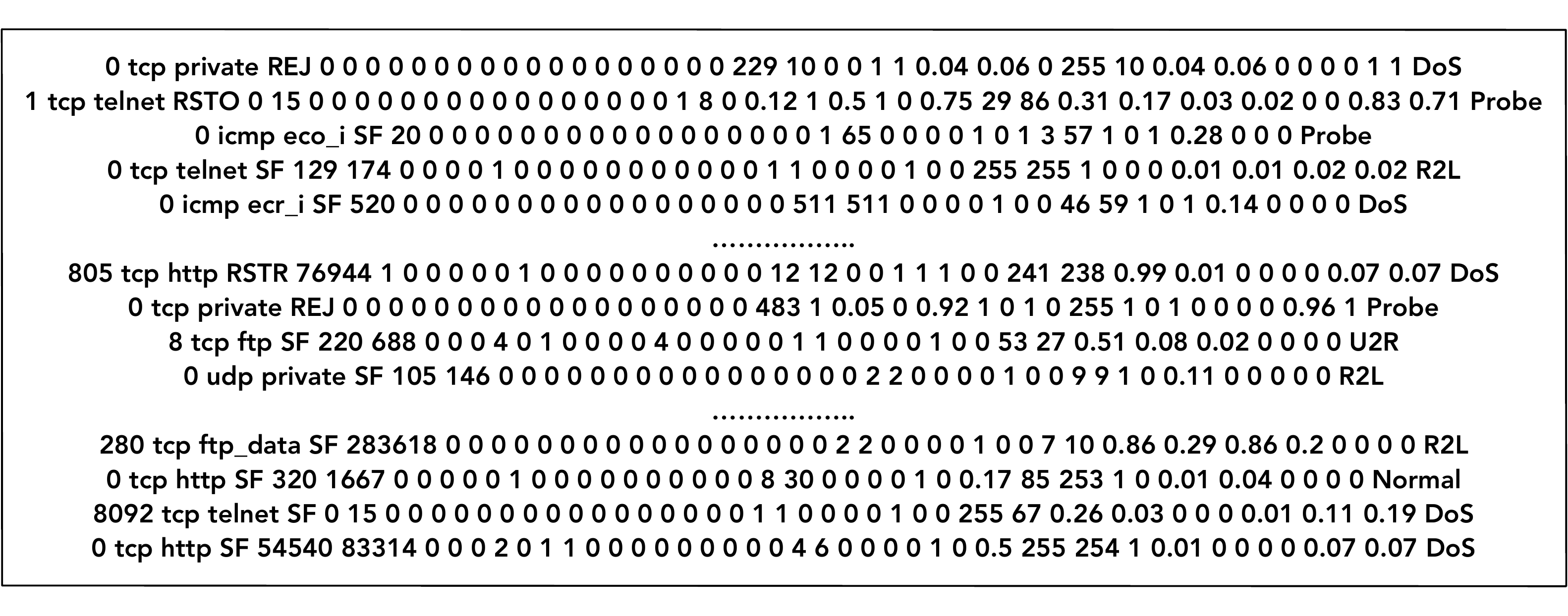}}
\caption{}
\end{subfigure}\\
\begin{subfigure}{\linewidth}
\centerline{\includegraphics[width=.95\linewidth]{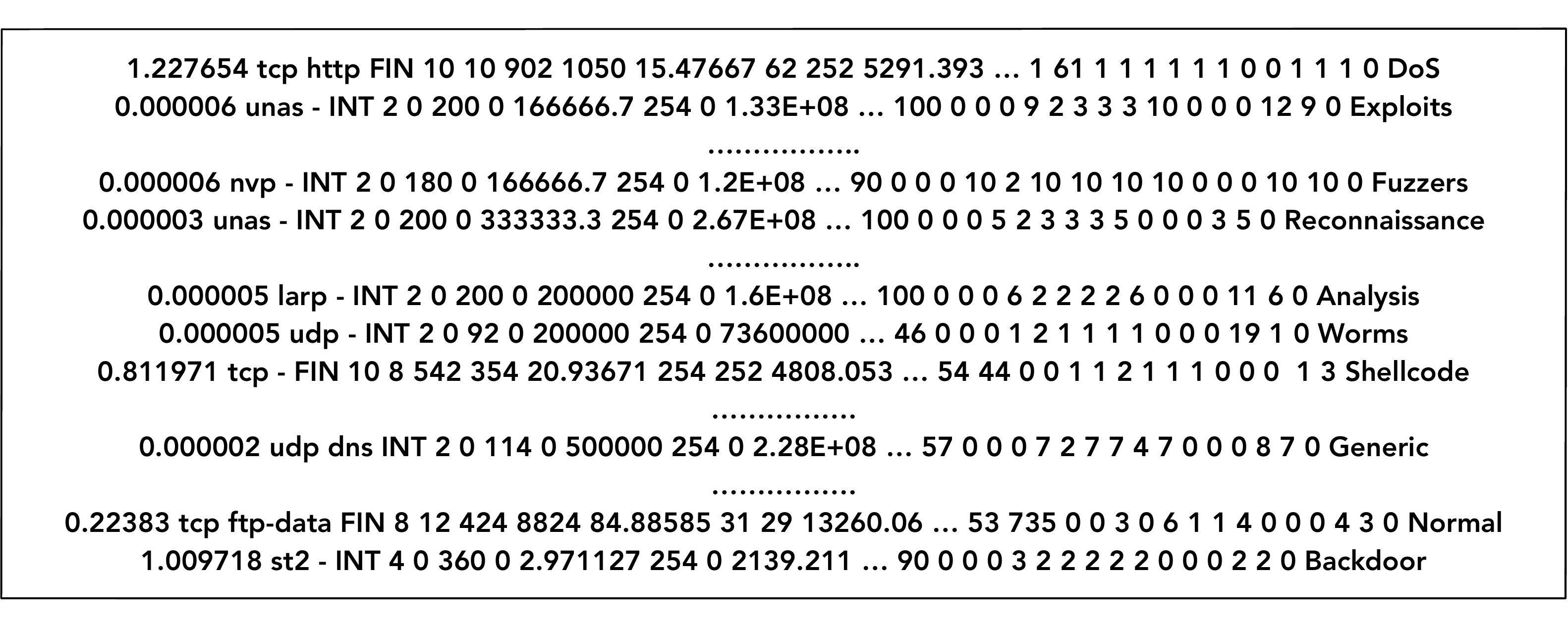}}
\caption{}
\end{subfigure}\\
\caption{Raw network traffic data. (a) NSL-KDD (41 original features), (b) UNSW-NB15 (42 original features).}
\label{fig:rawdata}
\end{figure}

\section{Evaluation}\label{evaluation}

The training environment used in our evaluation was based on TensorFlow backend, Keras and scikit-learn packages on a HP EliteDesk 800 G2 SFF Desktop with Intel (R) Core (TM) i5-6500 CPU @ 3.20 GHz processor and 16.0 GB RAM. 
The training was performed on two network intrusion datasets: NSL-KDD \cite{tavallaee2009detailed} and UNSW-NB15 \cite{moustafa2015unsw}.
Both datasets have removed a significant amount of redundant records from the originally collected data to ensure the trustworthiness of evaluation \cite{moustafa2016evaluation,mchugh2000testing}.
The NSL-KDD dataset consists of 5 categories: Normal, DoS, U2R, R2L and Probe, and the attack samples were collected based on a U.S air force network environment.
The UNSW-NB15 dataset includes 10 categories: Normal, DoS, Exploits, Generic, Shellcode, Reconnaissance, Backdoors, Worms, Analysis and Fuzzers, and the attack samples were collected from Common Vulnerabilities and Exposures\footnote{CVE: https://cve.mitre.org/}, Symantec\footnote{BID: https://www.securityfocus.com}, Microsoft Security Bulletin\footnote{MSD: https://docs.microsoft.com/en-us/security-updates/securitybulletins}. 

\begin{figure*}[t]
    \centering
    
    \begin{subfigure}[b]{.45\linewidth}
        \centering
        \includegraphics[width=\linewidth]{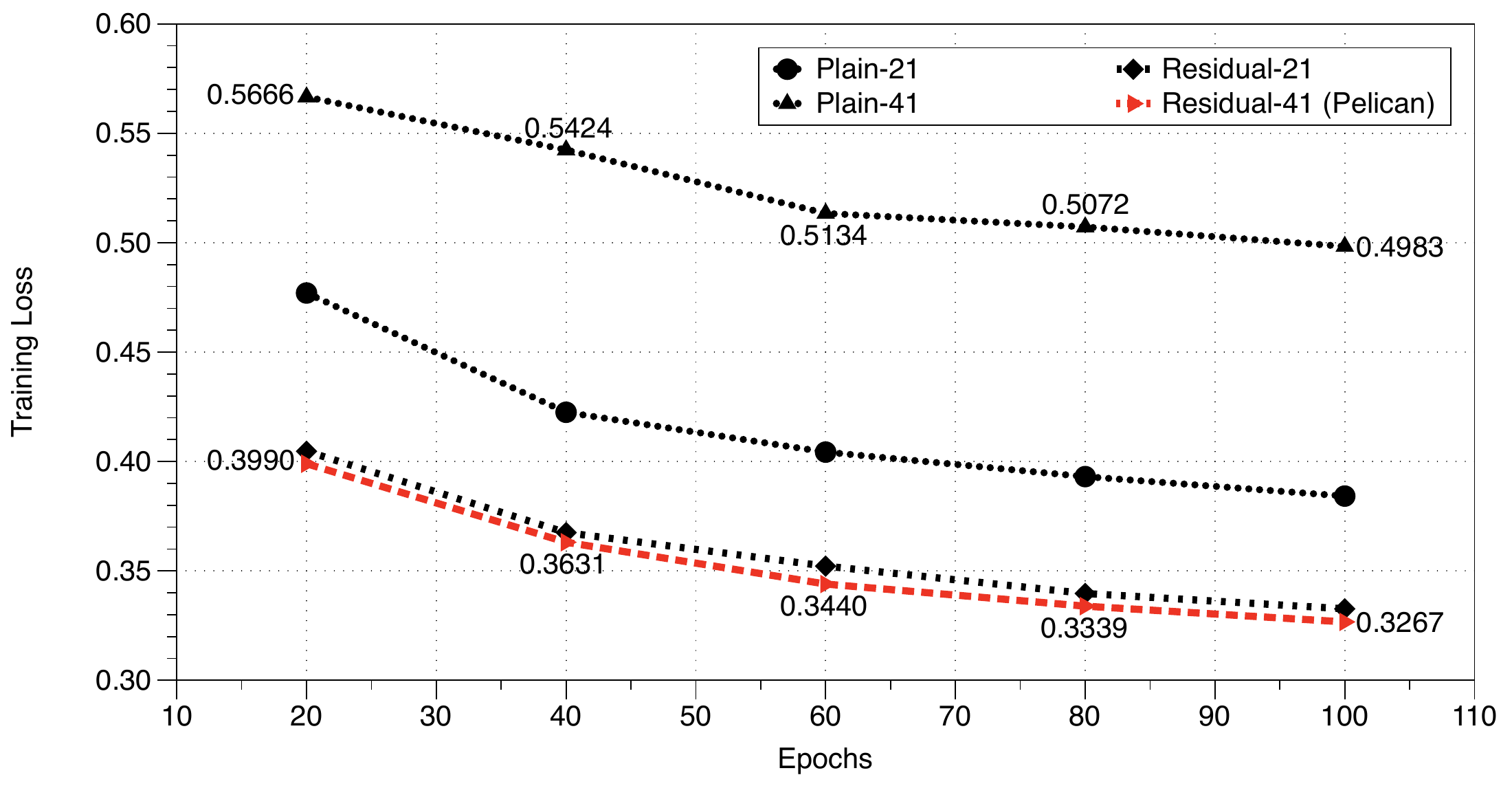}
        \caption{Training Loss on UNSW-NB15}
    \end{subfigure}%
 ~
    \begin{subfigure}[b]{.45\linewidth}
        \centering
        \includegraphics[width=\linewidth]{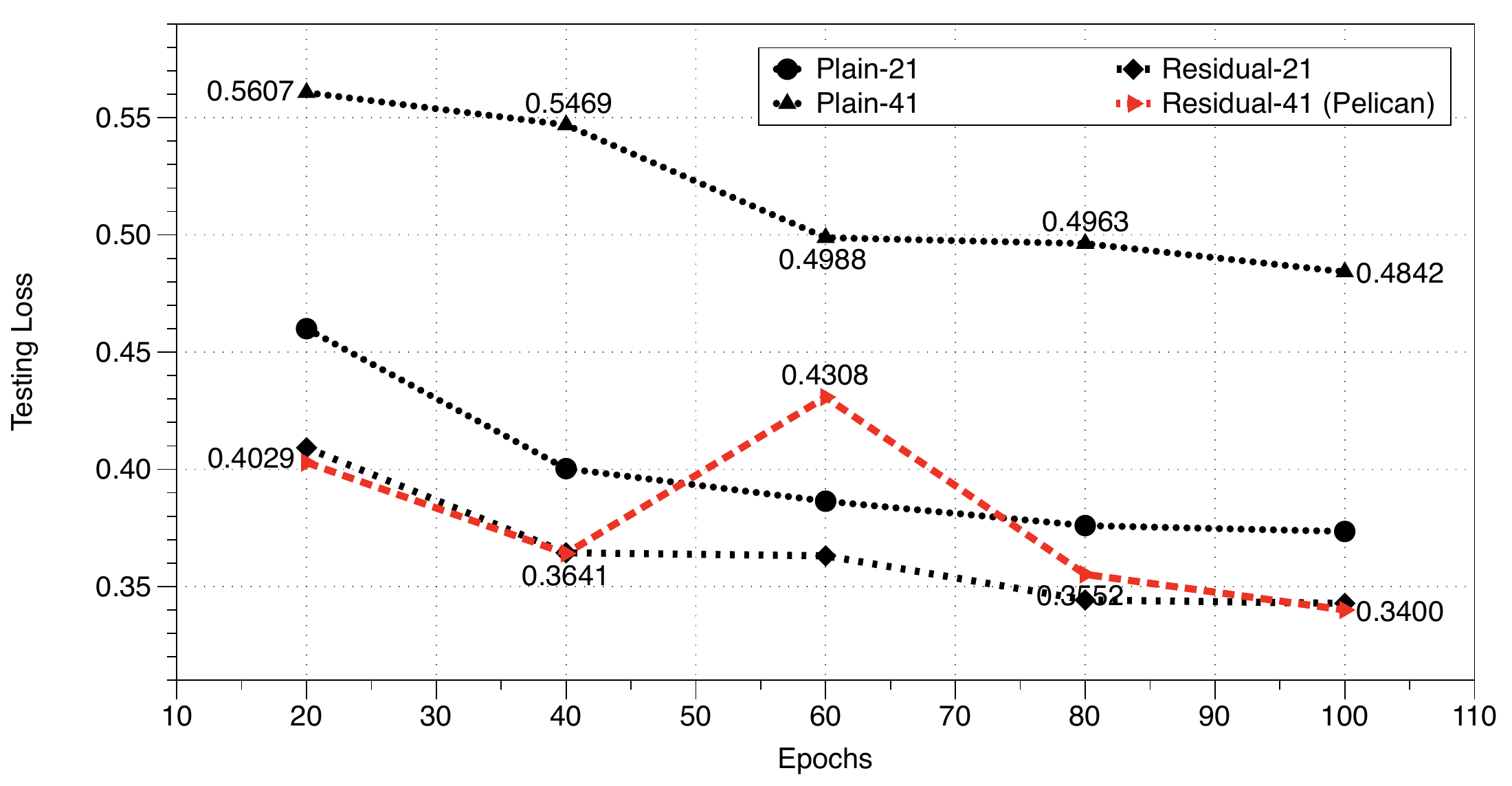}
        \caption{Testing Loss on UNSW-NB15}
    \end{subfigure}

    \begin{subfigure}[b]{.45\linewidth}
        \centering
        \includegraphics[width=\linewidth]{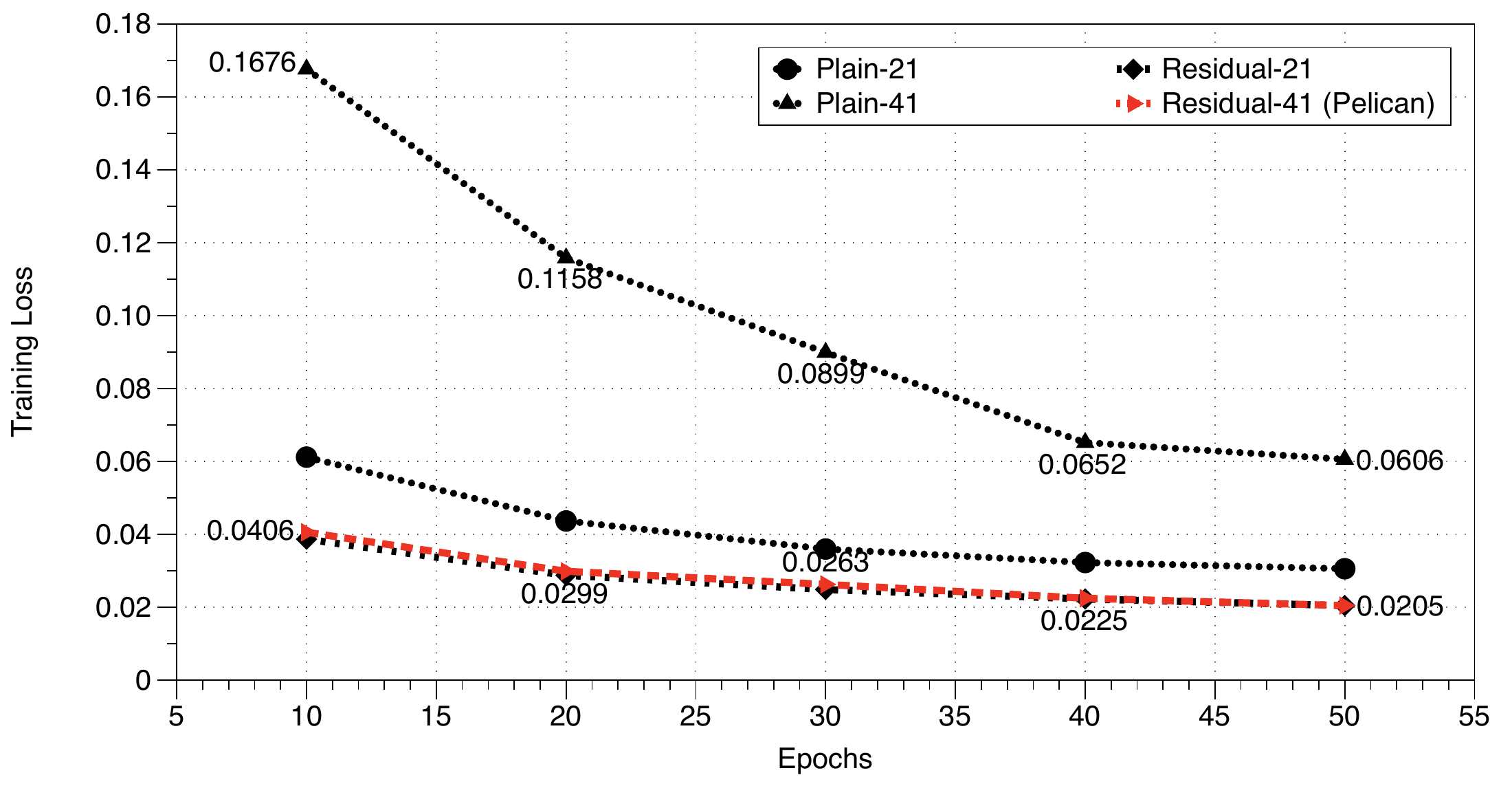}
        \caption{Training Loss on NSL-KDD}
    \end{subfigure}%
 ~
    \begin{subfigure}[b]{.45\linewidth}
        \centering
        \includegraphics[width=\linewidth]{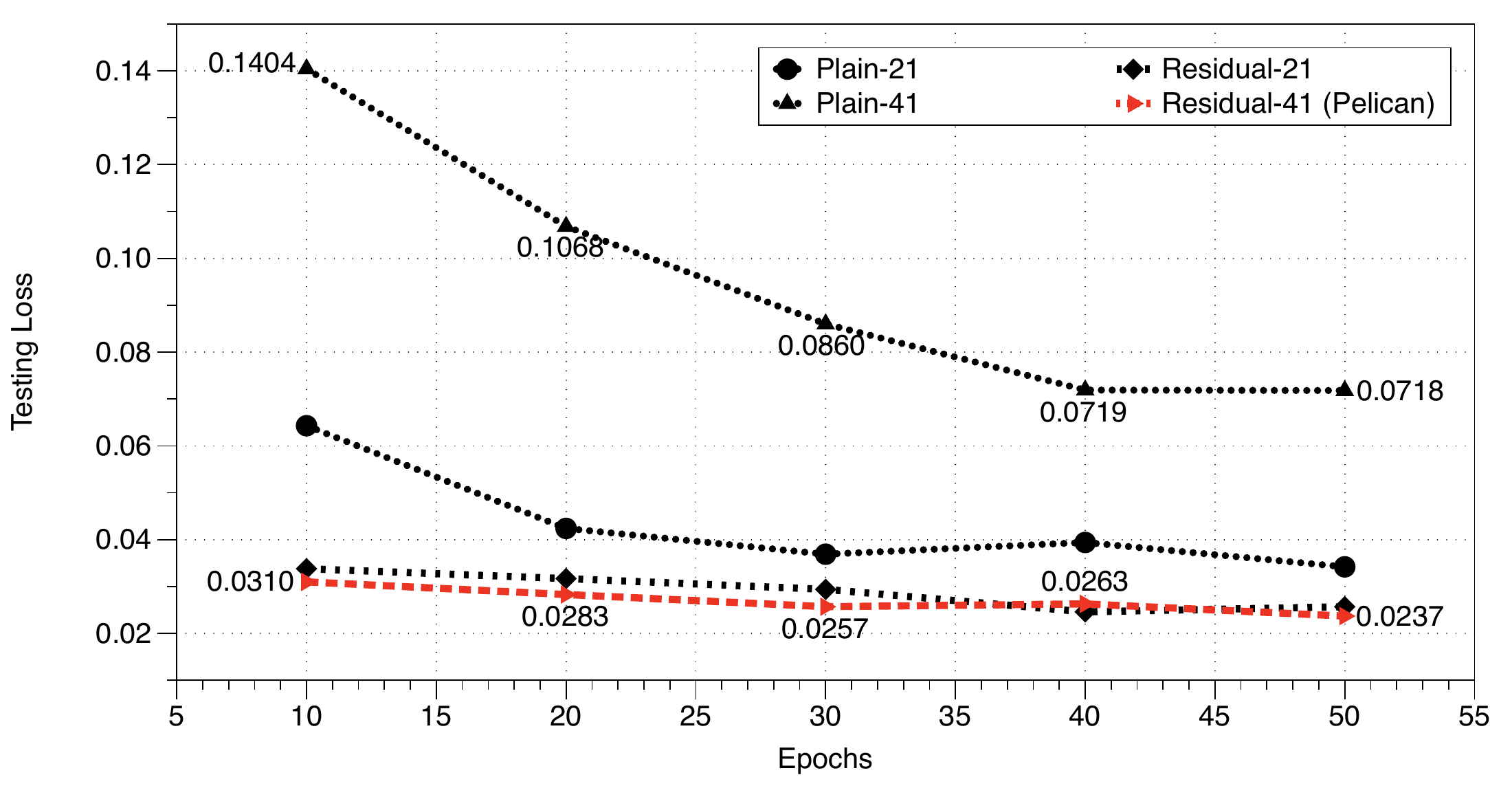}
        \caption{Testing Loss on NSL-KDD}
    \end{subfigure}
                \caption{A Comparison of Learning Performance of Four Tested Networks on UNSW-NB15 and NSL-KDD.}
                \label{fig.loss}
\end{figure*}

\subsection{Data Preprocessing}
There were 148,516 and 257,673 data records from NSL-KDD and UNSW-NB15 used in the evaluation.
Before training, we needed to preprocess the data, which consists of three steps:

\textbf{Step 1, Numerical Conversion}: Since the neural network could not recognize the textual notation, 
such as `tcp', `private' and `http' in the raw data as shown in Fig~\ref{fig:rawdata}, we converted them into numerical values.
Here, we used the `get\_dummies' function in Pandas \cite{mckinney-proc-scipy-2010} for the conversion. 

\textbf{Step 2, Normalization}: The data in the dataset may have various distributions with different means and derivations, which may make the neural network learning difficult. Hence, we applied standardization to scale them with a mean of 0 and a standard deviation of 1.

\textbf{Step 3, Training/Testing Dataset Creation}: To address the problem of data deficiency \cite{wu2019transfer}, we used k-fold cross-validation to ensure a good training-testing proportion.
With the k-fold validation, a dataset was split into k subsets, where k-1 subsets were combined for training and the rest one was used for testing. 
Here, we set k=10.

\subsection{Hyperparameter Setting for Tested Networks}

The hyperparameter setting of training those networks introduced in Section~\ref{Pelican} (Plain-21, Residual-21, Plain-41, and Residual-41 of Pelican) on the two datasets are given in Table~\ref{Parameter Setting}. All networks were trained with RMSprop gradient descent algorithm \cite{tieleman2017divide}.

Since residual learning uses the ``add" operation, the output dimension of filters (number of filters) and recurrent units must be equal to the input shape.
In our experiment, after the data preprocessing, the input to the network for UNSW-NB15 is encoded to 196 features, and for NSL-KDD is encoded to 121 features. Therefore, the input shapes for the two datasets are (1,196) and (1,121) respectively. 
Hence, we set 196 filters and 196 recurrent units to learn UNSW-NB15, and similarly 121 filters and 121 recurrent units to learn NSL-KDD.

\subsection{Evaluation Metric}

We used three metrics to evaluate the performance: validation accuracy (ACC), detection rate (DR) and false-alarm rate (FAR), as defined below.

\begin{equation}
{\small ACC = \dfrac{TP+TN}{TP+TN+FP+FN},}
\label{formula.ACC}
\end{equation}
\begin{equation}
{\small DR = \dfrac{TP}{TP+FN},}
\label{formula.DR}
\end{equation}
\begin{equation}
{\small FAR = \dfrac{FP}{FP+TN},}
\label{formula.FPR}
\end{equation}
where TP and TN are, respectively, the number of attacks and the number of normal traffic correctly classified; FP is the number of actual normal records mis-classified as attacks, and FN is the number of attacks incorrectly classified as normal traffic.

\renewcommand{\baselinestretch}{1.4}
\begin{table}[t]
    \centering
        \caption{\textsc{Hyperparameter Setting}}

    \begin{tabular}[\linewidth]{c||c|c}
    \hline
     Category & UNSW-NB15 & NSL-KDD  \\
    \hline
    \hline
      Filter size & 196 & 121 \\
    \hline
      Kernel size & 10 & 10\\
      \hline
      Recurrent unit & 196 & 121\\
      \hline
      Dropout rate & 0.6 & 0.6\\
      \hline
      Epochs & 100 & 50 \\
      \hline
      Learning rate &0.01&0.01\\
      \hline
      Batch size & 4000 & 4000\\

    \hline
    \end{tabular}
    \label{Parameter Setting}
\end{table}
\renewcommand{\baselinestretch}{1}

\subsection{Training Loss and Testing Loss}\label{loss}

We recorded the training histories and compared the training and testing losses to examine how residual learning can mitigate the degradation problem in deep neural networks.
The training losses of the four networks on the two datasets are plotted in Fig.~\ref{fig.loss} (a)-(d).
As shown in Fig.~\ref{fig.loss},  Plain-21 has less losses than Plain-41, which indicates that adding more learning layers leads to poorer performance. However, by using residual learning, the losses reduce greatly; for the networks of the same depth, the residual network has a much lower loss than the plain network. It can also be observed that the deeper network, Residual-41, in most cases, shows smaller losses than the shallow one, Residual-21. The only exception happened during testing the both residual networks on the UNSW-NB15 dataset, as shown in Fig.~\ref{fig.loss} (b), which could be explained due to overfitting (More discussion on this will follow).

\subsection{True Attacks Detected vs False Alarms}

Table~\ref{tab:correct_attacks_false_alarms} presents the total number of correctly detected attacks (TP) and the total number of false alarms (FP) generated by the four networks on the two datasets. As can be seen from the table, the residual-41 can detect more attacks and at the same time generate less false alarms than other three designs. 

\renewcommand{\baselinestretch}{1.4}
\begin{table}[t]
\centering
\caption{\textsc{Total True Attacks Detected and Total False Alarms}}
\begin{tabular}[\linewidth]{c||c|c|c|c|c}
\hline
Dataset	&	&	Plain-21	&	Residual-21	&	Plain-41	&	Residual-41	\\
	\hline \hline
\multirow{2}{*}{NSL-KDD}	&	TP	&	14688	&	14702	&	14607	&	14732	\\
\cline{2-6}
	&	FP	&	62	&	58	&	52	&	50	\\
	\hline
\multirow{2}{*}{UNSW-NB15}	&	TP	&	22094	&	22265	&	21211	&	22321	\\
\cline{2-6}
	&	FP	&	220	&	136	&	399	&	121	\\
\hline
\end{tabular}
\label{tab:correct_attacks_false_alarms}
\end{table}
\renewcommand{\baselinestretch}{1}

\renewcommand{\baselinestretch}{1.4}
\begin{table}[t]
\centering
\caption{\textsc{Testing Performance on NSL-KDD}}
\begin{tabular}[\linewidth]{c||c|c|c}
\hline
Struture & DR\%& ACC\%& FAR\%\\
\hline
\hline
Plain-21 &98.70 & 98.92 & 0.80\\
\hline
Plain-41 & 97.56 & 98.37 & 0.67\\
\hline
Residual-21 & 98.81 & 99.01 & 0.73\\
\hline
\textbf{Residual-41 (Pelican)} & \textbf{99.13} & \textbf{99.21} & \textbf{0.65}\\
\hline
\end{tabular}
\label{dr-far-nslkdd}
\end{table}
\renewcommand{\baselinestretch}{1}

\renewcommand{\baselinestretch}{1.4}
\begin{table}[t]
\centering
\caption{\textsc{Testing Performance on UNSW-NB15}}
\begin{tabular}[\linewidth]{c||c|c|c}
\hline
Structure & DR\%& ACC\%& FAR\%\\
\hline
\hline
Plain-21 & 97.42 & 85.76 & 2.37\\
\hline
Plain-41 & 93.73 & 82.33 & 4.29\\
\hline
Residual-21 & 97.86 & 86.42 & 1.46\\
\hline
\textbf{Residual-41 (Pelican)} & \textbf{97.75} & \textbf{86.64} & \textbf{1.30}\\
\hline
\end{tabular}
\label{dr-far-unsw}
\end{table}
\renewcommand{\baselinestretch}{1}

\subsection{Overall Performance}

Table~\ref{dr-far-nslkdd} and Table~\ref{dr-far-unsw} shows the detection rate (DR), validation accuracy (ACC), and false alarm rate (FAR) of the four networks tested on the NSL-KDD and UNSW-NB15 datasets, respectively. From the two tables, we can conclude that:
\begin{itemize}
    \item The residual networks outperform the plain networks with a high detection rate, a good validation accuracy and a low false alarm rate.
    \item With the increasing number of learning layers, a deeper residual network can achieve better performance than a shallow residual network.
\end{itemize}

\subsection{Limitations on Experiments}\label{limitation}

Though our evaluation here has basically demonstrated the effectiveness of using residual learning for deep neural network for network intrusion detection, our experiment results are restricted by some limitations, and the training data insufficiency and the low capacity of computing resources are two major limiting factors we encountered, which are discussed below.

\subsubsection{Training Data Insufficiency}

Generally, a deep neural network requires sufficient data for it to learn effectively and be able to cater for large scale data. Insufficient training data may lead to overfitting, which can be partially addressed by dropout, as has been mentioned in Section~\ref{pelican}. 
In our experiment, even though we already set a high dropout (0.6) to overcome overfitting, it was still an issue, as manifested by our experiment results shown in Fig.~\ref{fig.loss} (b). 
This situation indicates that the deep network model needs more data to fit.
However, because of privacy and security concerns, a sufficient cyber-attack dataset is much expensive to obtain.
Currently, NSL-KDD and UNSW-NB15 are only two trustworthy cyber-attack datasets that are free of redundancy. We hope the training datasets of sufficiently large will be available in future so that Pelican can be further improved and evaluated.

\subsubsection{Low Capacity of Computing Resources}

Our Pelican model can be easily scaled up with more learning layers. However in our evaluation, we only managed to test on the networks of up to 41-parameter layers due to the limited computing resources we had. 
When we added more layers, the training became extremely slow and laborious, which can be improved with more powerful computing devices.

\subsection{A Comparative Study }\label{techniques}

To further evaluate our Pelican DNN (of currently 41 layers), we compare it with a set of typical  machine learning based designs, as briefly described below, for  network intrusion detection.

\textbf{Support Vector Machine (SVM)} \cite{ahmad2018performance}: SVM is a classical machine learning approach that uses a kernel function, such as Gaussian kernel (RBF), to learn high-dimensional data. 
But as pointed out in \cite{bengio2006curse}, it has a low generation capability on learning large scale data. 

\textbf{Adaptive Boosting (AdaBoost)} \cite{hu2013online}: It is an ensemble learning approach that uses many cascaded weak classifiers (such as decision trees) to construct a stronger classifier to learn complex tasks.
The advantage of using many weak classifiers is its ability to mitigate the overfitting problem. 
However, AdaBoost often does not work well on imbalanced datasets.

\textbf{Random Forest (RF)} \cite{zhang2008random}: RF is also an ensemble learning approach. But compared to AdaBoost, it uses a different strategy of weight allocation.
Apart from having ability to reduce overfitting, RF can also handle imbalanced data. But its generalization capability often relies on the specification of features to be learned and
for the effectiveness of learning, a large number of features are required.

\textbf{MultiLayer Perceptron (MLP)} \cite{pal1992multilayer}: MLP is an early class of feed-forward neural network that uses Back-Propagation to learn non-linear problems.

\textbf{Convolution Neural Network (CNN, or ConvNet)} \cite{hinton2012neural}: CNN is the most popular deep neural network that has been used for image recognition and has gained great successes.
With the help of convolution operation, CNN has an ability to generate spatial representations from raw data.

\textbf{Long Short Term Memory (LSTM)} \cite{hochreiter1997long}: LSTM is a recurrent neural network. By generating temporal representations from learning, LSTM has been successfully applied to speech recognition and machine translation. 
LSTM is similar to GRU we used in our residual block but LSTM has a higher computing cost \cite{dey2017gate}.

\textbf{HAST-IDS} \cite{wang2017hast}: It is a recently proposed intrusion detection system that uses a tandem CNN+LSTM model (first learning spatial representations by CNN, then learning temporal representations by LSTM) as the core decision strategy.

\textbf{LuNet} \cite{peilun}: It is also an CNN+LSTM based intrusion detection system, similar to HAST-IDS. But LuNet uses a different architecture for effective learning of both spatial and temporal features of the input data and shows a better performance than HAST-IDS. 

We trained each of those designs on UNSW-NB15 dataset. Their detection rate (DR), accuracy (ACC) and false alarm rate (FAR) are given in Table~\ref{compare_techniques}.

\renewcommand{\baselinestretch}{1.4}
\begin{table}[t]
\centering
\caption{\textsc{A Comparison of Pelican's Performance with Classical Techniques (Based on UNSW-NB15)}}

\begin{tabular}[\linewidth]{c||c|c|c}
\hline
Design & DR\%& ACC\%& FAR\%\\
\hline
\hline
 AdaBoost & 91.13 & 73.19 & 22.11\\
\hline
 SVM (RBF) &83.71 & 74.80 & 7.73\\
\hline
 HAST-IDS & 93.65 & 80.03 & 9.60\\
\hline
 CNN &92.28 & 82.13 & 3.84\\
\hline
 LSTM &92.76 & 82.40 & 3.63\\
\hline
 MLP & 96.74 &84.00 & 3.66\\
\hline
 RF & 92.24 & 84.59 & 3.01\\
\hline
 LuNet & 97.43 & 85.35 & 2.89\\
\hline
\textbf{Pelican} & \textbf{97.75} & \textbf{86.64} & \textbf{1.30}\\
\hline
\end{tabular}
\label{compare_techniques}
\end{table}
\renewcommand{\baselinestretch}{1}

As can be seen from the table, among all the designs examined, Pelican shows the best performance - with the highest detection rate and accuracy, and the lowest false alarm rate - which further demonstrates the effectiveness of our design with residual learning for network intrusion detection.

\section{Background}\label{background}

In this section, we provide some background knowledge related to network intrusion detection 
and then present our two point of views for contemporary NIDS designs:
1) using anomaly detection is not suitable for real time intrusion detection for large scale network and 2) using supervised learning is effective but challenging. 

The idea of network intrusion detection systems (NIDS) is proposed to improve the capability of security incidents response\cite{denning1987intrusion,mukherjee1994network}.
By monitoring the network activities, NIDS can timely alert the suspicious and malicious network behaviour to network administrators.

Security experts believe that most of new attacks are variants of known attacks. The traditional NIDS designs prevent malicious behaviours by using specific attack signatures from the known attack libraries and have developed some successful products, such as Snort \cite{roesch1999snort} and Zeek-IDS \cite{paxson1999bro}.
However, the signature-based solution lacks of intelligence to discover advanced variants of previously known attacks. Hence, two alternative strategies have been proposed to address the problem: supervised learning and anomaly detection.

The principle of anomaly detection is that we only need to learn a profile of normal traffic whereas the outliers are considered as attacks.
Statistical learning \cite{moustafa2017novel,tan2014detection,tan2014detection,tsai2010triangle,saurabh2016efficient} and unsupervised learning \cite{portnoy2000intrusion,casas2012unsupervised,zanero2004unsupervised} are currently primary techniques used for anomaly detection.
The anomaly detection for network intrusion detection is not very suitable and practical for the reasons below: 

\textbf{Reason one} \cite{sommer2010outside}: Anomaly detection often leads to a high false alarm rate. However, a secure and dependable NIDS system not only should have zero tolerance of attacks but also should not block legitimate requests.

\textbf{Reason two} \cite{garcia2009anomaly}: Even if we can reduce false alarm rate by, for example, developing very sophisticated statistical or learning algorithm, a current notation of normal profile may not be fully representative in the future due to highly evolved network. Therefore, anomaly detection may only work well in a controlled network and may be less useful and practical in most of commercial network environments.

In contrast to anomaly detection, supervised learning requires a well-defined threat model to learn the underlying distinctions between normal and abnormal behaviour (such as Pelican and other classical supervised machine learning techniques shown in Table~\ref{compare_techniques}).
Supervised learning often produces a lower false alarm rate (FAR) and has more stable performance than anomaly detection. 
Hence, it is currently considered as a more reasonable and practical approach than anomaly detection for network intrusion detection.
However, there are some challenges that need to be addressed.

\textbf{Challenge one}: Supervised learning requires a good and continuously updated threat model to maintain its intelligence for detecting more new attacks.
However, due to the privacy and information security concerns, network attack data are often much expensive to collect. 

\textbf{Challenge two}: Even if supervised learning can contain FAR better than anomaly detection, the false alarm rate is still not satisfactory enough compared to human experts. 
Hence, more effective detection approaches are still required to further reduce the false alarm rate.

\section{Conclusion}\label{conclusion}
In the paper, we have examined the performance degradation problem in deep neural networks and investigated how the problem can be alleviated by using residual learning.

We presented a deep residual network, Pelican, for network intrusion detection, and compared its performance with traditional plain networks and its shallower residual network.

Our work has shown that residual learning not only can be used with CNN for image classification as has been demonstrated in the literature, but also can be used with CNN+RNN structures for network intrusion detection.
Compared to a set of state-of-the-art classical machine learning based techniques, by using residual learning, the deep neural network, Pelican, has a high capability for network intrusion detection.

It must be stressed that our evaluation is not thorough due to the limitations we encountered (small training datasets and lack of powerful computing facilities). A deeper Pelican with more learning layers will be investigated in the future when large training datasets and powerful computing resources become available.

\bibliographystyle{ieeetr}
\bibliography{sample-base}

\end{document}